\documentclass[runningheads]{llncs}
\usepackage{graphicx}

\usepackage[misc]{ifsym}
\usepackage[strings]{underscore}
\begin{document}
\title{A Genetic Feature Selection Based Two-stream Neural Network for Anger Veracity Recognition }
%
%
\author{Chaoxing Huang\thanks{This work was done when Chaoxing was at ANU} {\Letter} , \inst{}
Xuanying Zhu, \inst{}
Tom Gedeon \inst{}}
\authorrunning{Chaoxing Huang et al. }
%
\institute{Research School of Computer Science,\\Australian National University\\ 
ACT 2601 AUSTRALIA \\
\email{first name.last name@anu.edu.au}}
\maketitle              
\tocauthor{Chaoxing Huang, Xuanying Zhu, Tom Gedeon}
\toctitle{A Genetic Feature Selection Based Two-stream Neural Network for Anger Veracity Recognition}
\begin{abstract}
People can manipulate emotion expressions when interacting with others. For example, acted anger can be expressed when the stimulus is not genuinely angry with an aim to manipulate the observer. In this paper, we aim to examine if the veracity of anger can be recognized from observers’ pupillary data with computational approaches. We use Genetic-based Feature Selection (GFS) methods to select time-series pupillary features of  observers who see acted and genuine anger as video stimuli. We then use the selected features to train a simple fully connected neural network and a two-stream neural network. Our results show that the two-stream architecture is able to achieve a promising recognition result with an accuracy of 93.6\% when the pupillary responses from both eyes are available. It also shows that genetic algorithm based feature selection method can effectively improve the classification accuracy by 3.1\%. We hope our work could help current research such as human machine interaction and psychology studies that require emotion recognition.

\keywords{Anger veracity  \and Two-stream architecture \and Neural network \and Genetic algorithm}
\end{abstract}
\section{Introduction}
\paragraph{}The veracity of emotions plays an essential role in human interaction. It influences people's view towards others after the observer observes a certain emotion\cite{1}.  In reality,  human beings are sometimes very poor at telling whether a person's emotion is genuine or posed, especially in the scenario that humans are usually asked to use verbal information to make the prediction\cite{21}.  This kind of mistake may negatively affect some current research like psychological studies which includes emotion observation\cite{22}. Thus,  it is worth looking into the problem of using computational algorithms to take the physiological responses of humans to aid the recognition. Also, in human-machine interaction, it is important to let the machine know whether a human's emotion is disguised or genuine if the interaction involves emotion\cite{3}. There has been work using physiological signals of observers who are exposed to emotional stimuli to interpret the emotion of the stimuli. In \cite{2} and \cite{3}, a classifier is trained to identify if a person's smile or anger is genuine or posed. Meanwhile, a human thermal data based algorithm is proposed to analyse human's stress in \cite{4}.

Neural Networks (NNs) are able to learn their parameters automatically via back-propagation  and can be used to map physiological data to emotion veracity. However, since human beings interact with the environment, it is likely that physiological data being collected by  sensors are  noisy.  Noisy features can dampen the learning process of the NN on the data-set, since the model needs to  learn the underlying pattern of the noise. On the other hand, the model can overfit the dataset when the training time has to be escalated. Therefore, it is crucial to look into the problem of selecting useful features from the physiological data-set. Anomaly detection has a profound studied history. One of the most classic methods is the generative model learning approach \cite{12}. However, this method requires a cumbersome learning process and relies heavily on distribution assumptions. Other works have also been done to detect noisy and fraud features \cite{6,7}, and genetic algorithm is used in \cite{5} to select features without much human intervention. Since the noise in physiological data is usually not obvious to non-expert humans and has long temporal sequences, the evolutionary based genetic algorithm becomes a reasonable way for avoiding intractable manual selection. In the original work of \cite{3}, it is shown that using pupillary data, the model can recognise anger veracity with an accuracy of 95\%, which is a significant improvement over verbal data. However, the collected data may contain environment-affected noise and the sensors occasionally fail to collect physiological data at some time stages. When we take the time-series information into consideration, not all the recorded data from the sensor plays essential roles in classification due to this noisiness and redundancy, which thus requires feature selection, and the genetic algorithm provides a way to achieve this. Therefore, we study the effect of  genetic-based feature selection (GFS) \cite{15} on anger veracity recognition in this paper. \par
The contribution of this paper is two-fold:
\begin{itemize}
\item We adopt a two-stream neural-network to effectively use the physiological data (pupil diameters) from the two eyes of humans to predict the anger veracity of the emotion stimuli. 
\item We adopt the genetic-based feature selection method to select useful features from the noisy temporal data due to environment noise and occasional sensor failures (e.g. eye-blinks) and thus to enhance the recognition performance. 
\end{itemize}
In this paper, we first tune a baseline NN with one hidden layer by taking the pupillary data from one eye as input.  Then we apply GFS to the time-series data and verify our proposed two-stream model can handle the binocular pupillary information. The rest of this paper is organised as follows: Section 2 introduces the NN architecture, and the GFS pipeline. Section 3 is about the experiments and  results. Discussions are also provided in this section. Section 4 includes future work and concludes this paper.
\section{Method}
\subsection{Dataset}
We use Chen et al's anger dataset\cite{3}. The dataset was collected by displaying 20 video segments to 22 different persons (observers).  The observers watched the presenters' anger expression in the video and the   pupillary response of the observers were collected by an eye-tracking sensor. A sample in the dataset means the pupillary data of a person which was collected when the person watched a video, and each of them is labelled with "Genuine" or "Posed", meaning a genuine anger expression or a posed anger expression is observed. The videos have various lengths and the recorded data sequence length of each samples varies from 60 time-steps to 186 time-steps. The sensor recording rate for each of the samples is 60Hz. The dataset contains the pupillary response from each observer's two eyes at each time-step as well as the mean statistics.

\subsection{Network Architecture}
\subsubsection{Baseline Architecture}
In the baseline model, we adopt a simple fully-connected neural network architecture with one hidden layer with $n$ hidden neurons.  There are three potential choices of activation function in our NN, which are Sigmoid, Tanh and ReLU. We will investigate the effect of different choices of $n$ and activation function type in the experiment part. Since this is a binary classification, we choose cross-entropy loss as the loss function.\par

\subsection{Two-stream architecture}
Inspired by the two-stream architecture in video recognition \cite{13,14}, we adopt a two-stream fully connected architecture in our classification task, which is shown in Figure 1. For every stream, the sub-stream network is the baseline model and the feature vector from the two streams are fused together to a one-layer fully connected layer for final prediction. There are two potential kinds of input to the network. The first scenario is, the first stream takes the pupillary temporal data from the left eye and the second stream takes the pupillary data from the right eye. The second scenario is, the first stream takes the pupillary temporal data from the left (right) eye and the second stream takes the pupillary diameter differences at each time step from the left (right) eye. The pupillary difference for each time step is  the data at the current time step minus the data at the previous step. 
\begin{figure}[h]
 \centering
\includegraphics[scale=0.8]{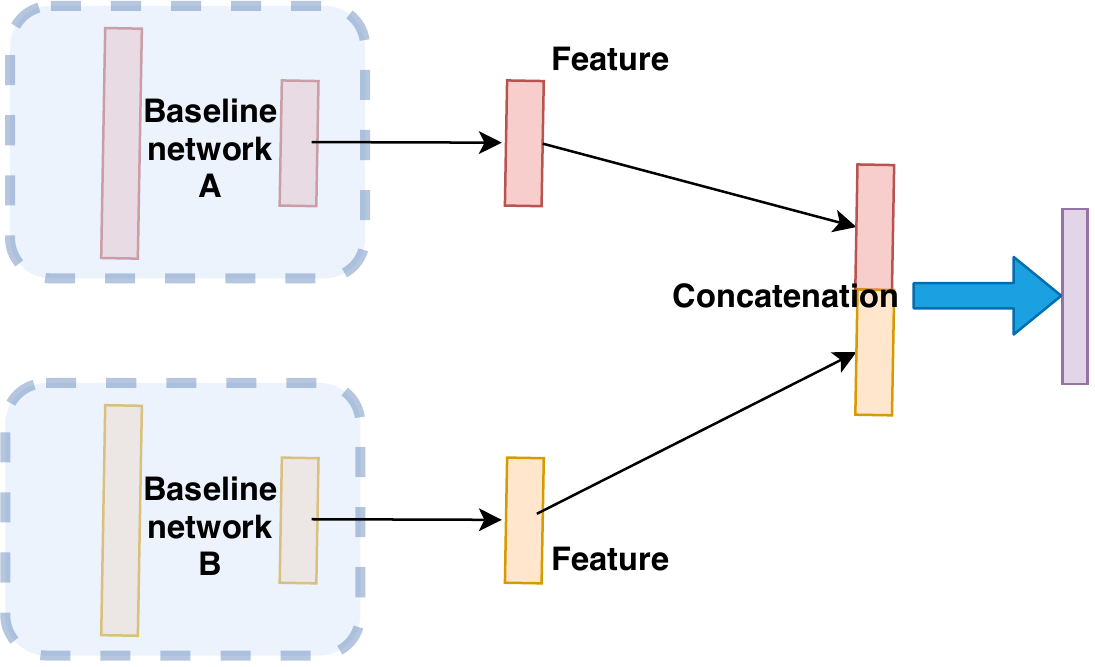}
\caption{Two-stream network architecture}
\label{fig1}
\end{figure}
\subsection{Data Pre-processing and Feature Selection}
\subsubsection{Data pre-processing}

 The data in the dataset is temporal, with values from both eyes at each time time-step. We regard every sequence of time-series data of each sample as an input vector to the neural network. To deal with the length varying issue, we use zero padding to pad every feature vector to the same length ($186\times 1$).
\subsubsection{Genetic-based feature selection}
The feature selection mask is indicated by a binary vector with the length of the feature-vector (0 for omitting a feature and 1 for keeping a feature). In a genetic algorithm, the selection mask is regarded as the chromosome. We first initialize the population size as $n+1$, and we adopt a neural network to compute the validation classification accuracy as the  fitness value. Note that for different chromosomes, the input size of the neural network is different, and thus we are not only doing a feature selection but also conducting a network architecture selection. We adopt a tournament-based reproduction\cite{18}, in which we create $\frac{n}{2}$ sets of tournament-group, and we randomly choose a fix size of members from the current generation to form the tournament-groups as the population pool for generating off-spring. Note that we actually repeat $\frac{n}{2}$ times of tournament group creation, which means one chromosome can appear in different tournament groups. In each tournament-group, two parents are selected by using the selection probability which is obtained by its normalized fitness value in the population (proportional selection). The crossover generates two off-springs by a one-point crossing. Therefore, the tournament-reproduction can generates $n$ off-springs, while the selected one is the chromosome with the highest fitness value in the current generation.   Every generated off-spring goes through a mutation process to increase the gene diversity.  To sum up, the population of each generation retains at $n+1$ while the parents' selection in every generation's reproduction need to go through a fierce tournament competition. The pipeline is shown in Figure 2. 

\begin{figure}[h]
 \centering
\includegraphics[width=12cm]{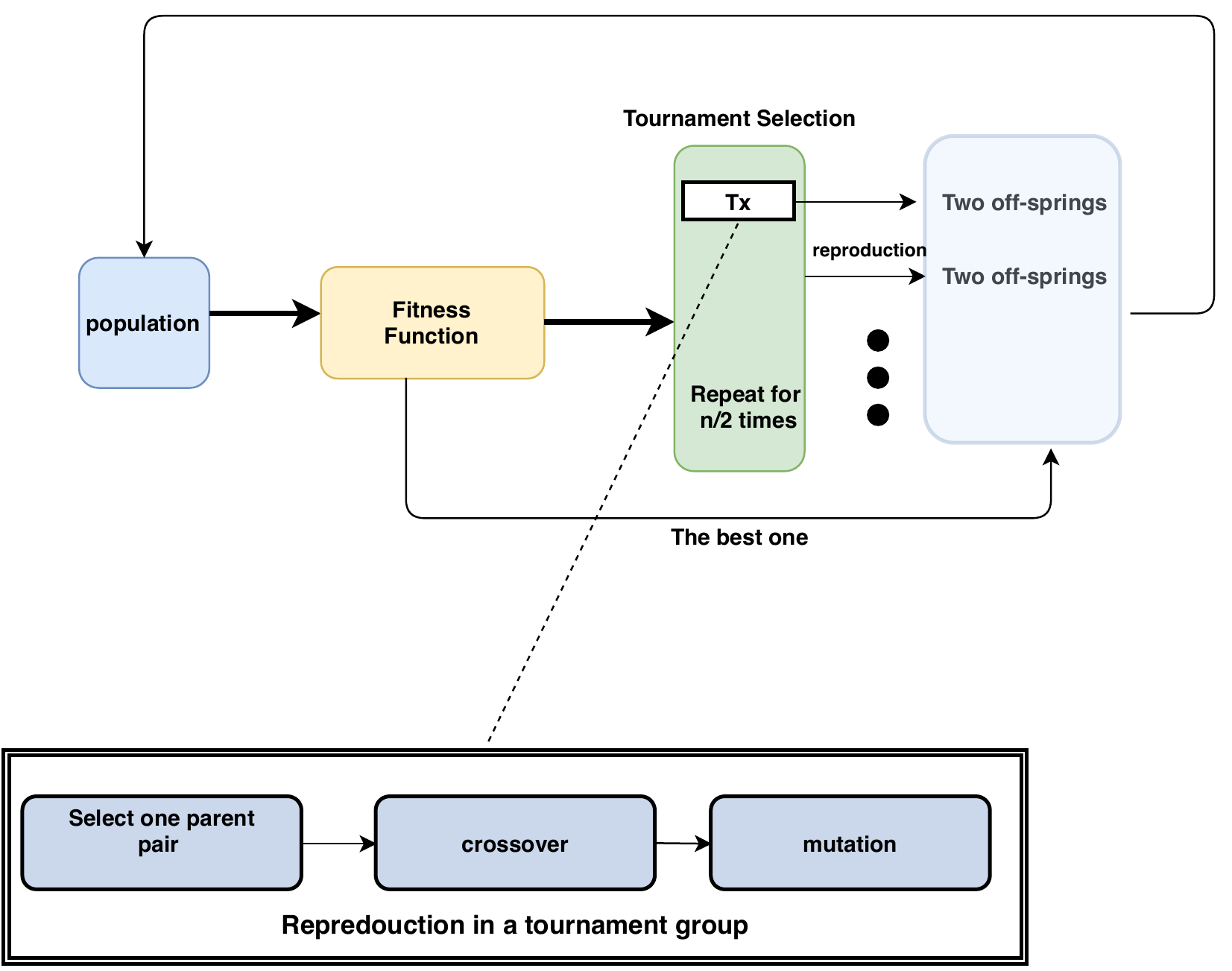}
\caption{Genetic-algorithm pipeline}
\label{fig1}
\end{figure}

\section{Experiments and Discussions}
\subsection{Experiment settings}
We first shuffle the data-set and randomly split out 80\% of the data as training patterns. The rest of the data are for testing. We use Python3.6 and  Pytorch 1.0 \cite{10} to implement the experiment and the environment is Windows 10. Since the training set is small, we only use an i7-8750H CPU for computation and we adopt batch gradient descent with an Adam \cite{9} optimizer. For the optimizer, the hyper-parameters are: $\beta_1=0.9$ and $\beta_2=0.999$. The learning rate is set to be 1e-4 and a weight-decay of 1e-5 is used to prevent over-fitting. The iteration number of training is set to be 1000 epochs. The random seed numbers are set as 1000, 2000, 3000, 4000 and 5000. We will denote them as sets 1 to 5 respectively in this context. All the results are reported based on five-fold cross-validation by using set 1 to set 5.\par

\subsection{Baseline model}
We train the model under different settings of hidden neuron number $n$ and different activation function types. We take the left eye pupillary data as input. The results, which are the average accuracy of the 5 runs, are shown in Table 1. As is clearly shown, the performance of using ReLU activation function with 60 neurons shows significant advantage over those of Tanh and Sigmoid since it avoids the problem of gradient diminishing and overfitting. In the rest of the experiments, we will use the model using ReLU with 60 hidden units as the baseline. We also present the precision, recall and F1 score of the baseline model after five runs in Table 2.  \par

\begin{table}
\centering
\caption{Test accuracy of the baseline model (\%)}\label{tab1}
\setlength{\tabcolsep}{10mm}{\begin{tabular}{||c c c c||} 
 \hline
 $n$ & \textbf{ReLU} & \textbf{Tanh} & \textbf{Sigmoid} \\ [0.5ex] 
 \hline\hline
 30 & 90.26 & 91.03 & 91.03 \\
 \hline
 40 & 91.54 & 91.79 & 92.05 \\
 \hline
 50 & 91.54 & 90.30 & 92.05 \\ 
 \hline
 60 & \textbf{93.08} & 91.54 & 91.80 \\ [1ex] 
  \hline
 70 & 90.52 & 91.27 & 91.80 \\
  \hline
 80 & 91.54& 91.28& 92.05 \\ [1ex] 
 \hline
\end{tabular}}
\end{table}

\begin{table}
\centering
\caption{Precision, recall and F1 score of baseline model}\label{tab1}
\setlength{\tabcolsep}{8mm}{\begin{tabular}{||c c c c||} 
 \hline
 \textbf{Metric} & \textbf{Genuine} & \textbf{Posed} & \textbf{Average}\\ [0.5ex] 
 \hline\hline
 Precision & 95.68 & 91.23 & 93.46 \\
 \hline
 Recall & 88.57 & 97.65 & 93.11 \\
 \hline
 F1 score & 91.99 & 94.33 &93.16 \\ 
 [1ex] 
 \hline
\end{tabular}}
\end{table}

\subsection{Experiments on GFS and two-stream architecture}
\subsubsection{Classification without feature-selection}
We first conduct experiments on training the fully-connected classifier without applying GFS on the temporal features. We compare the 5-set averaged results of different input settings of both single-stream and two-stream models, and the results are shown in Table 3. The hidden layer number is still 60. We also present the precision, recall as well as the F1 score of the best model here. The results are shown in Table 4. We can also notice that using information from two eyes achieves a better performance than only using information from either single eye, and this is aligned with our human intuition.  It should also be noted that taking the diameter differences into consideration does not provide an improvement, this  may be due to the fact that the diameter change in every time-step is very small and does not provide much significant temporal information. 

\begin{table}
\centering
\caption{Test accuracy of different model (\%)}\label{tab1}
\setlength{\tabcolsep}{10mm}{\begin{tabular}{||c c||} 
 \hline
 \textbf{Input} & \textbf{Test Accuracy} \\ [0.5ex] 
 \hline\hline
 Double-eyes (two-stream) & \textbf{93.58}  \\ 
 \hline
Left-eye (single-stream) & 93.08  \\
 \hline
 Right-eye (single-stream) & 92.31 \\
 \hline
 Left-eye+Left-differences (two-stream) & 91.03  \\
 \hline
 Right-eye+Right-differences (two-stream) & 91.03  \\  [1ex] 
 \hline
\end{tabular}}
\end{table}

\begin{table}
\centering
\caption{Precision, recall and F1 score of double-eye two-stream model}\label{tab1}
\setlength{\tabcolsep}{8mm}{\begin{tabular}{||c c c c||} 
 \hline
 \textbf{Metric} & \textbf{Genuine} & \textbf{Posed} & \textbf{Average}\\ [0.5ex] 
 \hline\hline
 Precision & 96.88 & 91.30 & 94.09 \\
 \hline
 Recall & 88.57 & 97.67 & 93.12 \\
 \hline
 F1 score & 92.54 & 94.38 &93.39 \\ 
 [1ex] 
 \hline
\end{tabular}}
\end{table}

\subsubsection{Effectiveness of GFS}
We study the effect of applying Genetic feature selection to the model. During the feature selection stage, 80\% of the training data is used for training while 20\% of the training data is used for validation to compute the fitness value. After selecting the best chromosome, we train our model on the entire training set. For every tournament group, the member number is set to be 9. The generation number is set to be 10. The population number is set to be 21 and the mutation rate is 0.001. In the two-stream case, both the streams use the same feature selection mask. We conduct the GFS experiment on the double-eye two stream model as well as the left-eye single stream model. The results are shown in Table 5. \par

\begin{table}
\centering
\caption{Performance of applying GFS}\label{tab1}
\setlength{\tabcolsep}{4mm}{\begin{tabular}{||c c c c c||} 
 \hline
  \textbf{Two-stream}  &\textbf{Genuine}&\textbf{Posed } & \textbf{Average} & Accuracy : 96.15 \% \\ [0.5ex] 
 \hline\hline
 Precision & 97.06 & 95.45 & 96.26 & \\ 
 \hline
 Recall & 94.29 & 97.67 & 95.98 & \\ 
 \hline
 F1 score & 95.65 & 96.54 & 96.10 & \\ 

 \hline
 \textbf{One-stream}  &\textbf{Genuine}&\textbf{Posed } & \textbf{Average} & Accuracy: 94.82 \%\\ [0.5ex] 
  \hline\hline
 Precision & 94.29 & 95.35 & 94.82 & \\ 
 \hline
 Recall & 94.29 & 95.35 & 94.82 & \\ 
 \hline
 F1 score & 94.29 & 95.35 & 94.82 & \\
 \hline
\end{tabular}}
\end{table}
As we can see from Table 3 and 5, by using the GFS method, the single-stream model with left eye input can achieve a better performance by applying feature selection, the single-stream model using feature selection  can even surpass the performance of the double-eye two stream model that does not have feature selection. Moreover, the double-eye two stream model with feature-selection achieves the best performance among all. The result reveals an underlying drawback, that using zero padding to fill up the length of those "short" vector create feature redundancy, and feature-selection can improve the recognition performance. 
\subsection{Discussion}
\subsubsection{Why does the GFS improve the recognition performance?}
Perhaps the most plausible explanation is that the GFS helps to eliminate some of the noisy pupillary data during collection. In the original paper using the dataset, it is mentioned that the eye-blinks of the humans exist, and which is prone to affect some of the collected pupillary diameter values. In fact, we have found that many of the feature selection masks have many zero entries before the zero padding stage, which means those actual collected data are removed. 

\section{Conclusion and Future Work}
 This paper examines applying two-stream neural networks with genetic-based feature selection on anger veracity recognition, to tackle the problem of noisy data collection and sensor failure during physiological data collection. From the experimental results, it can be concluded that the two-stream architecture can effectively handle the data from both eyes of humans and it is crucial to take these binocular physiological reactions into consideration when doing anger veracity recognition. It can also be concluded that applying genetic-based feature selection can effectively improve the model performance and remove redundant or noisy features. \par
In our work, we use the same feature mask for selecting features from both eyes. One might question that the data behaviour from two eyes may be different and applying an identical mask overlooks this possibility to a certain extent.  Indeed, the asymmetric reaction of both eyes can affect the data collection of the sensor differently, and it is worth looking into the problems of finding two feature selection masks that can reflect the linkage and difference between the two eyes.\par
As for the model that uses time-series data, our fully-connected model requires zero padding to deal with varying length data, which creates redundancy and reduces flexibility. Therefore, it is worth looking into the method of applying  RNN/LSTM \cite{19} or Transformer models \cite{20} in the future. \par

\section{Acknowledgments}
We would like to thank those contributors who collected and provided the dataset. We would also like to thank the Human Centred Computing team at ANU for providing useful advice for this paper.
%
%
%
%

\end{document}